\documentclass[english]{smfart}

\usepackage[english]{babel}
\usepackage{graphicx}
\usepackage{subfigure}
\usepackage{amsthm,amssymb,amsbsy,amsmath,amsfonts,amssymb,amscd}
\newcommand{\E}{\mathrm{E}}
\newcommand{\V}{\mathrm{var}}
\newcommand{\C}{\mathrm{cov}}
\newcommand{\p}{\mathrm{P}}

\newtheorem{prop}{Proposition}
\usepackage[usenames,dvipsnames]{pstricks}
\usepackage{dsfont}

\title[Additive Covariance kernels for high-dimensional GP modeling]{Additive Covariance kernels for high-dimensional Gaussian Process modeling}

\author{Nicolas Durrande}
\address{LSTI - Ecole Nationale Sup\'erieure des Mines de St-Etienne, 29 rue Ponchardier, 42023 St Etienne, France}
\email{durrande@gmail.com}

\author{David Ginsbourger}
\address{Institute of Mathematical Statistics and Actuarial Science, University of Berne, Alpeneggstrasse 22, 3012 Bern, Switzerland}
\email{david.ginsbourger@stat.unibe.ch}

\author{Olivier Roustant}
\address{LSTI - Ecole Nationale Sup\'erieure des Mines de St-Etienne, 29 rue Ponchardier, 42023 St Etienne, France}
\email{roustant@emse.fr}

\keywords{Additive Models, Kriging, Gaussian Processes, GAM, Interpretable Modeling, Computer Experiment}

\date{\today}

\begin{document}
\frontmatter

\begin{abstract}
Gaussian process models --also called Kriging models-- are often used as mathematical approximations of expensive experiments. However, the number of observation required for building an emulator becomes unrealistic when using classical covariance kernels when the dimension of input increases. In oder to get round the curse of dimensionality, a popular approach is to consider simplified models such as additive models. The ambition of the present work is to give an insight into covariance kernels that are well suited for building additive Kriging models and to describe some properties of the resulting models.
\end{abstract}

\begin{altabstract}
La mod\'elisation par processus gaussiens --aussi appel\'ee krigeage-- est souvent utilis\'ee pour obtenir une approximation math\'emathique d'une fonction dont l'\'evaluation est co\^uteuse. Cependant, le nombre d'\'evaluations n\'ecessaires pour construire un mod\`ele bas\'e sur des noyaux de covariance usuels devient d\'emesur\'e lorsque la dimension des variables d'entr\'ee augmente. Afin de contourner le fl\'eau de la dimension, une alternative bien connue est de consid\'erer des mod\`eles simplifi\'es comme les mod\`eles additifs. Nous pr\'esentons ici une classe de noyaux de covariance adapt\'ee \`a la construction de mod\`eles de krigeage additifs et nous d\'ecrivons certaines propri\'et\'e des mod\`eles obtenus.
\end{altabstract}

\maketitle

\mainmatter
\section{Introduction}

The study of numerical simulators often deals with calculation intensive computer codes. This cost implies that the number of evaluations of the numerical simulator is limited and thus many methods such as uncertainty propagation, sensitivity analysis, or global optimization are unaffordable. A well known approach to circumvent time limitations is to replace the numerical simulator by a mathematical approximation called metamodel (but also emulator, response surface or surrogate model) based on the responses of the simulator for a limited number of inputs called the Design of Experiments (DoE). There is a large number of metamodels types and among the most popular we can cite regression, splines, neural networks. In this article, we focus on a particular type of metamodel: the Kriging method, more recently referred to as Gaussian Process modeling \cite{Rasmussen2006}. 
Originally presented in spatial statistics \cite{Cressie1993} as an optimal linear unbiased predictor of random processes, Kriging has become very popular in machine learning, where its interpretation is usually restricted to the convenient framework of Gaussian Processes (GP). The latter points of view allows the explicit derivation of conditional probability distributions for the response values at any point or set of points in the input space. 

\medskip

Since Kriging is usually based on local basis functions, it requires an increasing number of points in the DoE to cover the domain $D$ when the number of dimensions $d$ of the input space $D\subset \mathds{R}^{d}$ becomes high~\cite{Santner2003,K-T.Fang2006}. 
An approach to get around this issue is to consider specific features lowering complexity such as the family of Additive Models (AM). In this case, the emulator $m$ can be decomposed as a sum of univariate functions:
\begin{equation}
m(x) = \mu + \sum_{i=1}^{d}{m_i(x_i)},
\label{eq:AM}
\end{equation}
where $\mu \in \mathds{R}$ and the $m_i$'s may be non-linear. Since their introduction by Stones in 1985 \cite{stone1985additive}, many methods have been proposed for the estimation of additive models. We can cite the method of marginal integration \cite{newey1994kernel} and a very popular method described by Hastie and Tibshirani in \cite{Buja1989,Hastie1990}: the GAM backfitting algorithm. However, those methods do not consider the probabilistic framework of GP modeling and do not usually provide additional information such as the prediction variance. Combining the high-dimensional advantages of AMs with the versatility of GPs is the main goal of the present work. For the study functions that contain an additive part plus a limited number of interactions, details can be found found in a recent article of T. Muehlenstaedt~\cite{Muhl}.\medskip

The first part of this paper focuses on the unsuitability of usual separable kernels (e.g. power exponential and Mat\'ern) for high-dimensional modeling. The second part deals with additive Gaussian Processes, their associated kernels and the properties of associated Additive Kriging Models (AKM).
Finally, AKM is compared with standard Kriging models on a well known test function: the Sobol's g-function \cite{Saltelli2000}. It is shown within the latter example that AKM outperforms standard Kriging and produce similar performances as GAM. Due to its approximation performance and its built-in probabilistic framework, the proposed AKM appears as a serious and promising challenger for high-dimensional modeling.

\section{Towards additive Kriging}

\subsection{Additive random processes}

Lets first introduce the mathematical construction of an additive GP. A function $f:D \subset \mathds{R}^d \rightarrow \mathds{R}$ is additive when it can be written $f(x) = \sum_{i=1}^{d}{f_i(x_i)}$, where $x_i$ is the $i$-th component of the $d$-dimensional input vector $x$ and the $f_i$'s are arbitrary univariate functions. Let us first consider two independent real-valued Gaussian processes $Z_1$ and $Z_2$ defined over the same probability space $(\Omega, \mathcal{F},P)$ and indexed by $\mathds{R}$, so that their trajectories $Z_i(\cdot;\omega): t \in \mathds{R} \rightarrow Z_i(t;\omega)$ are univariate real-valued functions. 
Let $K_i:\mathds{R}\times\mathds{R} \rightarrow \mathds{R}$ be their respective covariance kernels and $\mu_1, \mu_2 \in \mathds{R}$ their means. 
Then, the process $Z$ defined over $(\Omega, \mathcal{F},P)$ and indexed by $\mathds{R}^{2}$, characterized by
\begin{equation}
\forall \omega \in \Omega \ \forall x \in \mathds{R}^{2} \ Z(x;\omega)=Z_{1}(x_1;\omega)+Z_{2}(x_2;\omega),
\label{eq_sum_proc}
\end{equation} 
\noindent
clearly has additive paths and has mean $\mu = \mu_1 + \mu_2$ and kernel $K(x,y)=K_1(x_1,y_1)+K_2(x_2,y_2)$. 
In this document, we call additive any kernel of the form $K:(x,y)\in \mathds{R}^d \times \mathds{R}^d \rightarrow K(x,y)=\sum_{i=1}^{d}{K_i(x_i,y_i)}$ where the $K_i$'s are symmetric positive-semidefinite (s.p.) kernels over $\mathds{R}\times\mathds{R}$. Although not commonly encountered in practice, it is well known that such a combination of s.p. kernels is also a s.p. kernel \cite{Rasmussen2006,gaetan2009spatial}. Moreover, one can show that the paths of any random process with additive kernel are additive in a certain sens:

\begin{prop}
\label{addproc}
Any (square integrable) random process $Z_{x}$ possessing an additive kernel is additive up to a modification. 
In essence, it means that there exists a process $A_{x}$ which paths are all additive, and such that $\forall x \in D,\ \mathrm{P}(Z_{x}=A_{x})=1$.
\end{prop}
The proof of this property is given in appendix for $d=2$. For $d=n$ the proof follows the same pattern but the notations are more cumbersome. 
Note that the class of additive processes is not actually limited to processes with additive kernels. For example, let us consider $Z_1$ and $Z_2$ two correlated Gaussian processes on $(\Omega, \mathcal{F},P)$ such that the couple $(Z_1,Z_2)$ is Gaussian. Then $Z_{1}(x_1)+Z_{2}(x_2)$ is also a Gaussian process with additive paths but its kernel is not additive. However, the term additive process will always refer to GP with additive kernels in this article. 

\subsection{Invertibility of covariance matrices}
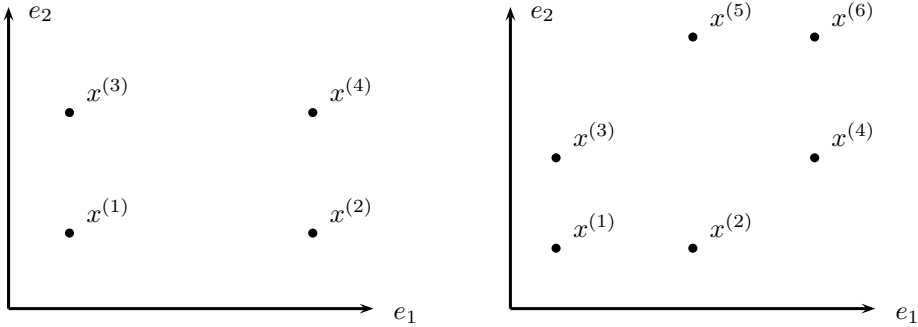
\begin{figure}[!ht]%
\begin{center}

\scalebox{1} 
{
\begin{pspicture}(0,-2.208125)(12.301875,2.208125)
\psline[linewidth=0.04cm,arrowsize=0.05291667cm 2.0,arrowlength=1.4,arrowinset=0.4]{->}(0.0,-1.8903126)(4.8,-1.8903126)
\psline[linewidth=0.04cm,arrowsize=0.05291667cm 2.0,arrowlength=1.4,arrowinset=0.4]{->}(0.0,-1.8903126)(0.0,2.1096876)
\psdots[dotsize=0.12](0.8,-0.8903125)
\psdots[dotsize=0.12](4.0,-0.8903125)
\psdots[dotsize=0.12](4.0,0.7096875)
\psdots[dotsize=0.12](0.8,0.7096875)
\usefont{T1}{ptm}{m}{n}
\rput(5.2214065,-1.9803125){$e_1$}
\usefont{T1}{ptm}{m}{n}
\rput(0.42140624,2.0196874){$e_2$}
\usefont{T1}{ptm}{m}{n}
\rput(1.3114063,-0.5803125){$x^{(1)}$}
\usefont{T1}{ptm}{m}{n}
\rput(4.5114064,-0.5803125){$x^{(2)}$}
\usefont{T1}{ptm}{m}{n}
\rput(1.3114063,1.0196875){$x^{(3)}$}
\usefont{T1}{ptm}{m}{n}
\rput(4.5114064,1.0196875){$x^{(4)}$}
\psline[linewidth=0.04cm,arrowsize=0.05291667cm 2.0,arrowlength=1.4,arrowinset=0.4]{->}(6.6,-1.8903126)(11.4,-1.8903126)
\psline[linewidth=0.04cm,arrowsize=0.05291667cm 2.0,arrowlength=1.4,arrowinset=0.4]{->}(6.6,-1.8903126)(6.6,2.1096876)
\psdots[dotsize=0.12](7.2,-1.0903125)
\psdots[dotsize=0.12](9.0,-1.0903125)
\psdots[dotsize=0.12](10.6,0.1096875)
\psdots[dotsize=0.12](7.2,0.1096875)
\usefont{T1}{ptm}{m}{n}
\rput(11.821406,-1.9803125){$e_1$}
\usefont{T1}{ptm}{m}{n}
\rput(7.021406,2.0196874){$e_2$}
\usefont{T1}{ptm}{m}{n}
\rput(7.711406,-0.7803125){$x^{(1)}$}
\usefont{T1}{ptm}{m}{n}
\rput(9.511406,-0.7803125){$x^{(2)}$}
\usefont{T1}{ptm}{m}{n}
\rput(7.711406,0.4196875){$x^{(3)}$}
\usefont{T1}{ptm}{m}{n}
\rput(11.111406,0.4196875){$x^{(4)}$}
\psdots[dotsize=0.12](10.6,1.7096875)
\usefont{T1}{ptm}{m}{n}
\rput(11.111406,2.0196874){$x^{(6)}$}
\psdots[dotsize=0.12](9.0,1.7096875)
\usefont{T1}{ptm}{m}{n}
\rput(9.511406,2.0196874){$x^{(5)}$}
\end{pspicture} 
}

\end{center}
\caption{2-dimensional examples of DoE which lead to non-invertible covariance matrix when using additive kernels. In both cases, one point can be removed from the DoE without any loss of information.}%
\label{fig:planprob}
\end{figure}

As mentioned in~\cite{chiles1999geostatistics} the covariance matrix $\mathrm{K}$ of the observations of an additive process $Z$ at a design of experiments $\mathcal{X}=(x^{(1)}\ \dots\ x^{(n)})^T$ may not be invertible even if there is no redundant point in $\mathcal{X}$. Indeed, the additivity of $Z$ may introduce linear relationships (that hold almost surely) between the observed values of $Z$ and lead to the non invertibility of $\mathrm{K}$. Figure~\ref{fig:planprob} shows two examples of designs leading to a linear relationship between the observation. For the left panel, the additivity of $Z$ implies that $Z(x^{(4)}) = Z(x^{(2)})+Z(x^{(3)})-Z(x^{(1)})$ a.s. so there is a linear relationship between the columns of $\mathrm{K}$ : $K(x^{(i)},x^{(2)})+K(x^{(i)},x^{(3)})-K(x^{(i)},x^{(1)})-K(x^{(i)},x^{(4)}) =0$ and therefore the matrix is not invertible. 


An approach which is in accordance with the aim of parsimonious evaluations of the simulator is to remove some points of the DoE in order to avoid any linear combination. Algebraic methods may be used for determining the subset of points leading to a the linear relationship. Indeed, the linear combination is given by the eigenvectors associated with the null eigenvalues, so the subset of points leading to the non invertibility of the covariance matrix can be obtained easily. However, the study of a procedure allowing to put aside unnecessary training points is out of the scope of this paper.

\subsection{Additive Kriging}
Let $f : D \rightarrow \mathds{R}$ be the function of interest (a numerical simulator for example), where $D \subset \mathds{R}^d$. The responses of $f$ at the DoE $\mathcal{X}$ are noted $F=(f(x^{(1)}) \ ... \ f(x^{(n)}))^T$. Simple Kriging relies on the hypothesis that $f$ is one path of a centered random process $Z$ with kernel $K$. The expression of the best predictor (also called Kriging mean) and of the prediction variance are:
\begin{equation}
\begin{split}
m(x) & =  \E \left[ \left. Z(x) \right| Z(\mathcal{X})=F  \right] = k(x)^T \mathrm{K} ^{-1}F \\
v(x) & =  \V \left[ \left. Z(x) \right| Z(\mathcal{X})=F  \right] = K(x,x) - k(x)^T \mathrm{K} ^{-1}k(x)
\end{split}
\label{eq:BP}
\end{equation}
where $k(\cdot)= \left( K(\cdot,x^{(1)})\ \dots\ K(\cdot,x^{(n)}) \right)^T$ and $\mathrm{K}$ is the covariance matrix of general term $\mathrm{K}_{i,j}=K(x^{(i)},x^{(j)})$. 
Note that these equations respectively correspond to the conditional expectation and variance in the case of a GP with known kernel.
In practice, the structure of $K$ is supposed to be known (e.g. power-exponential or Mat\'ern families) but its parameters are unknown. A common way to estimate them is to maximize the likelihood of $Z(\mathcal{X})=F$ \cite{Ginsbourger2009,Rasmussen2006}.
\medskip

\noindent
In some cases, the evaluation of $f$ includes an observation noise $\varepsilon$. To take this into account in the expression of $m$ and $v$ correspond to the conditional expectetion and variance of $Z$ knowing $Z(\mathcal{X})+\varepsilon(\mathcal{X})=F$. If we assume that $\varepsilon$ is a Gaussian white noise process with variance $\tau^2$, we obtain:
\begin{equation}
\begin{split}
m(x) & =  \E \left[ \left. Z(x) \right| Z(\mathcal{X})+\varepsilon(\mathcal{X})=F  \right] = k(x)^T (\mathrm{K}+\tau^2 \mathrm{Id}) ^{-1}F \\
v(x) & =  \V \left[ \left. Z(x) \right| Z(\mathcal{X})+\varepsilon(\mathcal{X})=F  \right] = K(x,x) - k(x)^T (\mathrm{K}+\tau^2 \mathrm{Id})^{-1}k(x).
\end{split}
\label{eq:BPnoise}
\end{equation}
As we can see, the covariance matrix of $\varepsilon(\mathcal{X})$ appears in the expression of $m$ and $v$. As we will use later, this remak is still valid when $\varepsilon(\mathcal{X})$ is a centered Gaussian vector.
\medskip

\noindent
Equations \ref{eq:BP} and \ref{eq:BPnoise} are valid for any s.p. kernel, so they can be applied with additive kernels. In this case, the additivity of the kernel implies the additivity of the Kriging mean so $m$ can be split in a sum of univariate submodels $m_1,\dots,m_d$. For example in dimension 2 with additive kernel $K(x,y)=K_1(x_1,y_1)+K_2(x_2,y_2)$ we have
\begin{equation}
\begin{split}
m(x) & =  (k_1(x_1)+k_2(x_2))^T(\mathrm{K_1+K_2})^{-1}F\\
& =  k_1(x_1)^T(\mathrm{K_1+K_2})^{-1}F + k_2(x_2)^T(\mathrm{K_1+K_2})^{-1}F\\
& = m_1(x_1) + m_2(x_2).
\end{split}
\label{eq:split}
\end{equation}

\noindent
Another interesting property concerns the variance: $v$ can be null at points that do not belong to the DoE. Let us consider a two dimensional example where the DoE is composed of the 3 points represented on the left pannel of figure~\ref{fig:planprob}: $\mathcal{X}=\{x^{(1)}\ x^{(2)} \ x^{(3)}\}$. Direct calculation (see Appendix B) shows that the prediction variance at the point $x^{(4)}$ is equal to 0. 
This particularity follows from the fact that given the observations at $\mathcal{X}$ the value of the additive process at the point $x^{(4)}$ is known almost surely. In the next section, we illustrate the potential of AKM on an a toy example.

\subsection{Illustration and further consideration on a 2D example}
\begin{figure}[!ht]%
\begin{center}
\includegraphics[width=0.9\textwidth]{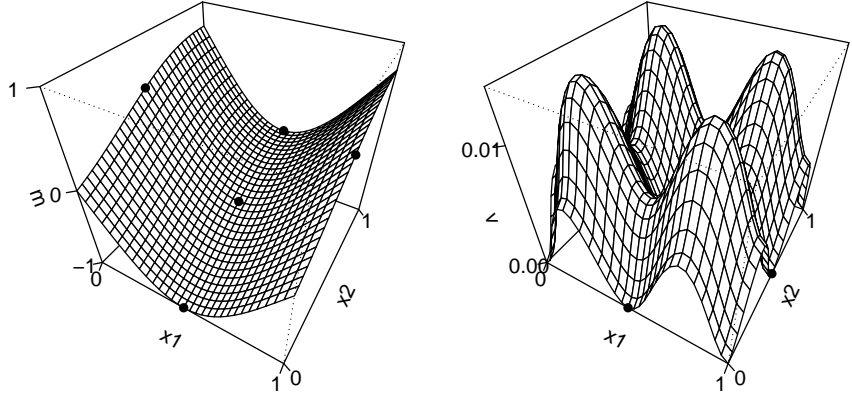}
\end{center}
\caption{Approximation of the function $f$ based on five observations (black dots). The left panel represents the best predictor and the right panel the prediction variance. The kernel here is the additive squared-exponential kernel with parameters $\sigma = (1\ 1)$ and $\theta = (0.6\ 0.6)$.}%
\label{fig:ex1dim2}
\end{figure}

We present here a first basic example of an additive Kriging model. We consider $D=[0,1]^2$, and a set of 5 points in $D$ where the value of the observations $F$ are arbitrarily chosen. Figure~\ref{fig:ex1dim2} shows the obtained Kriging model. We can see on this figure the properties we mentioned above: the Kriging mean is an additive function and the prediction variance can be null for points that do no belong to the DoE.

As we have seen in eq.~\ref{eq:split}, the expression of the first univariate model is
\begin{equation}
m_1(x_1)=k_1(x_1)^T(\mathrm{K_1}+\mathrm{K_2})^{-1}F.
\end{equation}
It appears that the effect of the direction 2 can be seen as an observation noise. We thus get the following expression for the prediction variance
\begin{equation}
v_1(x_1)=K_1(x_1,x_1)-k_1(x_1)^T(\mathrm{K_1}+\mathrm{K_2})^{-1}k_1(x_1).
\end{equation}

\begin{figure}[!ht]%
\begin{center}
\includegraphics[width=0.9\textwidth]{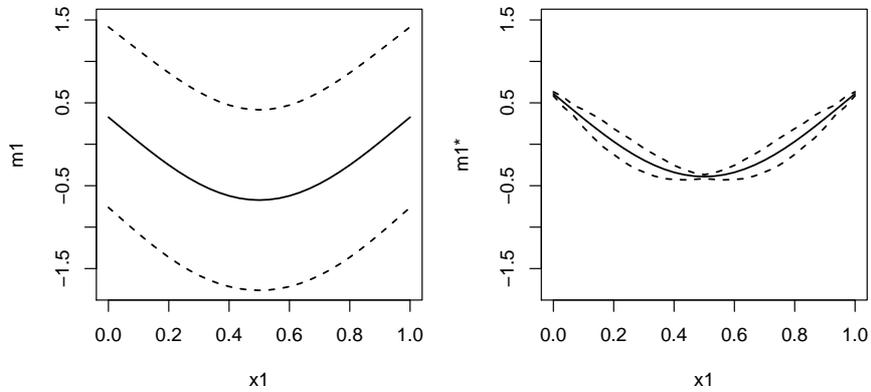}
\end{center}
\caption{Univariate models of the 2-dimensional example. The left panel plots $m_1$ and the 95\% confidence intervals $c_1(x_1)=m_1(x_1) \pm 2\sqrt{v_1(x_1)}$. The right panel shows the submodel of the centrated univariate effects $\tilde{m}_1$ and $\tilde{c}_1(x_1)=\tilde{m}_1(x_1) \pm 2\sqrt{\tilde{v}_1(x_1)}$}%
\label{fig:ex1dim2bis}
\end{figure}
The left panel of figure~\ref{fig:ex1dim2bis} shows the submodel $m_1$ and the associated $95\%$ confidence intervals. However, it appears that the confidence intervals are wide. This is because the submodels are define up to a constant. If we assume that $\int Z_i(s_i)\mathrm{d} s_i$ exist a.s.~\cite{fortet1985operateurs}, we can get rid of the effect of such a translation by emulating $Z_i(x_i)-\int{Z_i(s_i)\mathrm{d} s_i}$ conditionally to the observations:
\begin{equation}
\begin{split}
\tilde{m}_i(x_i)&=\E \left[ \left.Z_i(x_i)-\int{Z_i(s_i)\mathrm{d} s_i} \right| Z(\mathcal{X})=F  \right] \\
\tilde{m}_i(x_i)&=\V \left[ \left.Z_i(x_i)-\int{Z_i(s_i)\mathrm{d} s_i} \right| Z(\mathcal{X})=F  \right]
\end{split}
\end{equation}
The expression of $\tilde{m}_i(x_i)$ is straightforward whereas $\tilde{v}_i(x_i)$ requires more calculations given in Appendix C.
\begin{equation}
\begin{split}
\tilde{m}_i(x_i)&=m_i(x_i)-\int{m_i(s_i)\mathrm{d} s_i} \\
\tilde{v}_i(x_i)&= v_i(x_i) - 2 \int{K_i(x_i,s_i)\mathrm{d} s_i} + 2\int{k_i(x_i)^T \mathrm{K}^{-1} k_i(s_i) \mathrm{d} s_i} \\
& \hspace{2cm} + \iint{K_i(s_i,t_i)\mathrm{d} s_i \mathrm{d} t_i} - \iint{k_i(t_i)^T \mathrm{K}^{-1} k_i(s_i) \mathrm{d} s_i  \mathrm{d} t_i}
\end{split}
\end{equation}
The benefits of using $\tilde{m}_i$ and $\tilde{v}_i$ and then to define the submodels up to a constant can be seen on the right panel of figure~\ref{fig:ex1dim2bis}. Furthermore, as the submodels $\tilde{m}_i$ are univariate and centered, they may give a good approximation of the main effects of $f$ with relevant confidence intervals. At the end, the probabilistic framework gives an insight on the error for the metamodel but also for each submodel. 

\section{Kriging, high-dimensional input space and linear budget} 
\label{sec:khd}
We will see in this section that additive Kriging models can outperform usual Kriging models when the dimension of the input space becomes large. The notion of high-dimensional input space can be interpreted differently depending on the context. In our case, we will consider that an input space is high-dimensional when its dimension is larger than 10 and we will consider examples up to dimension 50. This exclude simulators for which one of the input is a picture or a map (for example groundwater flow simulators depending on permeability and porosity maps) where it is not unusual to deal with $50000$-dimensional input spaces.

\medskip
Most of the time, kernels used in computer experiment are power exponential or Mat\'ern kernels~\cite{Rasmussen2006}. For those kernels and for all other stationary kernels such that $\lim_{||x-y|| \rightarrow + \infty} K(x,y) = 0$, an observation at a point $x_1$ of the DoE has only a local influence on the emulator. This implies that the number of points required for modeling accurately a function increases exponentially with the dimension $d$ of the input space. However, large training sets are rather inconsistent with the context of emulating a costly-to-evaluate function and in contrast, a total budget of $10 \times d$ evaluations is sometimes advocated~\cite{loeppky2009choosing}. We now illustrate with an example that usual separable kernels are not appropriate for emulating high-dimensional functions for this budget of evaluation whereas additive kernels can advantageously be used to extract an additive trend. 

\medskip
Let $Z$ be a centered Gaussian process over $[0,1]^d$ with unit variance and an isotropic squared-exponential kernel 
\begin{equation}
K(x,y)= \prod_{i=1}^d \exp \left( - \frac{(x_i - y_i)^2}{\theta^2} \right).
\label{eq:noytest}
\end{equation}
Let $\mathcal{X}$ be a LH design of size $10 \times d$. Our aim is to investigate the reduction of variance obtained by conditioning $Z$ with respect to the observations $\mathcal{X}$ when $d$ increases. In order to quantify the proportion of variance explained by the emulator, we consider a test set $\mathcal{X}_t= \left( x_t^{(1)}, \dots,x_t^{(n_t)} \right) $ drawn from uniform distribution and we compute the following criterion
\begin{equation}
P = 1 - \frac{ \sum_{i=1}^{n_t} \V \left( \E \left(Z (x_t^{(i)} ) | Z(\mathcal{X}) \right) \right)}{ \sum_{i=1}^{n_t} \V \left( Z ( x_t^{(i)} ) \right) }.
\label{eq:P}
\end{equation}
According to the law of total variance, we have for all $i$ 
\begin{equation}
\V \left( Z ( x_t^{(i)} ) \right) = \V \left( \E \left(Z (x_t^{(i)} ) | Z(\mathcal{X}) \right) \right) + \E \left( \V \left(Z (x_t^{(i)} ) | Z(\mathcal{X}) \right) \right)
\label{eq:VT}
\end{equation}
so the values of $P$ are in $[0,1]$. As for a $Q_2$ criterion (see eq.~\ref{eq:Q2}), a value $P=1$ implies that $Z\left( x_t^{(i)} \right)$ is known a.s. for all test points whereas $P=0$ indicates that $\E \left( Z(\cdot) | Y(\mathcal{X}) \right) $ is no more predictive than $\E \left( Z(\cdot) \right)$. As $P$ do not take into account the distance between $m$ and the function to fit and as it priviledges overconfident models, this criteria is not ment to assess the quality of a GP emulator. However, it is well suited for studying the prediction ability of a GP emulator. 

\begin{figure}[!ht]%
\begin{center}
\includegraphics[width=0.6\textwidth]{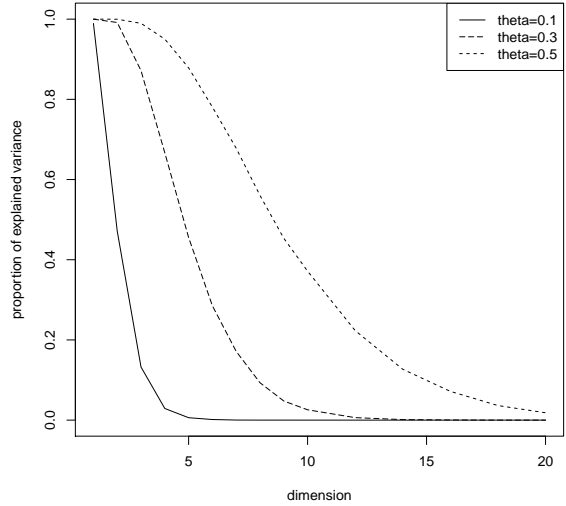}
\end{center}
\caption{Proportion of variance explained by $Z | Z(\mathcal{X})$ versus dimension. The $P$ criteria is computed for $n_t=10000$ test points uniformly distributed on $[0,1]^d$. The 3 curves correspond to different values of the range parameter $\theta$.}%
\label{fig:TPdinc}
\end{figure}

\medskip
As shown on figure~\ref{fig:TPdinc}, the proportion of explained variance collapses when the dimension increases, and this fall is all the more important as the range parameter $\theta$ is small. When the value of the range parameter $\theta$ is lower than half of the range of the data, simple or ordinary Kriging models with usual separable covariance are inappropriate to emulate high-dimensional functions for a budget of $10 \times d$ observations. However, further tests showed that such budget allows to build very predictive GP emulator up to $d = 100$ when $\theta=\sqrt{d}$.

\begin{figure}[!ht]%
\begin{center}
\includegraphics[width=0.6\textwidth]{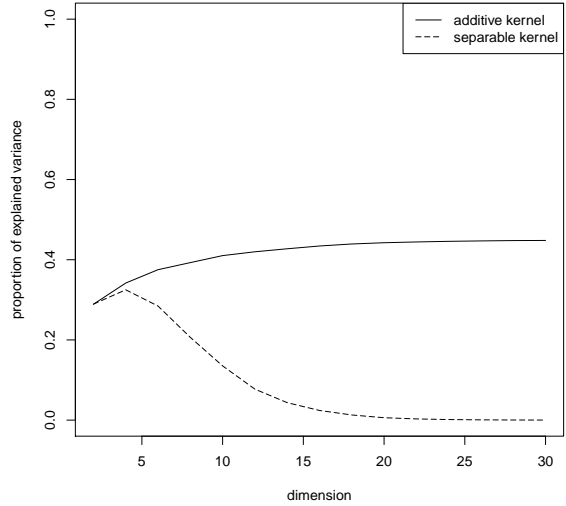}
\end{center}
\caption{Comparison of the predictivity of the approximation of $Y$ by $m_A$ and $m_S$.}%
\label{fig:TPdinc-AvsTP}
\end{figure}

\medskip
We will now consider a second example where the GP to be approximated has an additive component and compare the results of additive and non additive Kriging emulators. Let $Y_A$ and $Y_S$ be independant centered GPs indexed by $[0,1]^d$ with respectively an additive and a separable kernel: 
\begin{equation}
\begin{split}
K_A(x,y) &= \frac1d \sum_{i=1}^d \exp \left( - \frac{(x_i - y_i)^2}{0.5^2} \right) \\
K_S(x,y) &= \prod_{i=1}^d \exp \left( - \frac{(x_i - y_i)^2}{0.5^2} \right).
\end{split}
\label{eq:noytestAS}
\end{equation}
We define $Y$ as $Y = Y_A + Y_S$ so that the first half of the variance of $Y$ is explained by its additive part $Z_A$ and the second one by its separable part $Z_S$. We now compare the predictivity of 2 emulators:
\begin{equation}
\begin{split}
m_A(x) = \E (Y_A(x) | Y_A(\mathcal{X}) + Y_S(\mathcal{X}) ) = k_A(x)^t (\mathrm{K}_A + \mathrm{K}_S )^{-1} (Y_A(\mathcal{X}) + Y_S(\mathcal{X})) \\
m_S(x) = \E (Y_S(x) | Y_A(\mathcal{X}) + Y_S(\mathcal{X}) ) = k_S(x)^t (\mathrm{K}_A + \mathrm{K}_S )^{-1} (Y_A(\mathcal{X}) + Y_S(\mathcal{X})).
\end{split}
\label{eq:emtestAS}
\end{equation}
As we have seen previously, $m_A$ corresponds to the best predictor of an additive Kriging model with an observation noise given by $\mathrm{K}_S$. This emulator cannot explain the non additive part of $Y$. Reciprocally, $m_S$ is based on the separable kernel $K_S$ with an observation noise $\mathrm{K}_A$. This term may be able to cover both the additive and non additive part of $Y$ for a large number of observations. The prediction variance associated to those emulators is known analytically, so their predictivity can be compared as in the previous example. We observe on figure~\ref{fig:TPdinc-AvsTP} that the explained variance falls quickly to 0 when using a separable kernel whereas an emulator based on an additive kernel can capture efficiently the additive trend of the phenomena. On this example, and for a budget of $10 \times d$ evaluations, it appears that Kriging additive models clearly outperforms Kriging based on standard kernels.

\section{Application to the g-function of Sobol}
In order to illustrate the methodology and to compare it to existing algorithms, an analytical test case is considered. The function to approximate 
is the g-function of Sobol defined over $[0,1]^d$ by
\begin{equation}
g(x)= \prod_{k=1}^d \frac{|4x_k-2|+a_k}{1+a_k} \text{   with } a_k > 0
\label{eq:Gsob}
\end{equation}
This popular function in the literature \cite{Saltelli2000} is obviously not additive. However, depending on the coefficients $a_k$, $g$ can be very close to an additive function. As a rule, the g-function is all the more additive as the $a_k$ are large. One main advantage for our study is that the Sobol sensitivity indices can be obtained analytically so we can quantify the degree of additivity of the test function. For $i=1,\dots,d$ the indice $S_i$ associated to the variables $x_i$ is
\begin{equation}
S_i = \frac{\frac{1}{3(1+a_i)^2}}{\left[\prod_{k=1}^{d}1+\frac{1}{3(1+a_k)^2}\right]-1}.
\label{eq:SI}
\end{equation}
Here, we impose that the value of the parameters  $a_k$ is the same for all directions (ie $\forall k,\ a_k=a_1$). As the additivity of the g-function is tunable, we choose $a_1$ such that the variance of the additive part of $g$ correspond to $75 \%$ of the variance of $g$:
\begin{equation}
\sum_{i=1}^d S_i = 0.75  \Leftrightarrow d \frac{u}{(1+u)^d -1} = 0.75 \text{\quad with \quad} u = \frac{1}{3(1+a_1)^2}.
\label{eq:S8i}
\end{equation}
Eventually, the value of $a_1$ can be obtained by finding the zeros of a polynomial in $u$. Note that different values for $d$ leads to different values of $a_1$.

\medskip
For $d \in \{5,10,20,30\}$ and a Latin hypercube design based on $10 \times d$ points, we compare an Usual Kriging Model (UKM) with AKM and GAM. The two Kriging models are ordinary Kriging models since they include a constant term as a trend. As GAM is based on smoothing cubic splines, we choose a Mat\'ern $5/2$ kernel with observation noise for the Kriging models so as the different models have a similar regularity. The results for UKM and GAM are obtained with the DiceKriging~\cite{Roustant2009} and the GAM~\cite{Hastie2011}  \textit{R} packages available on the CRAN~\cite{RR}. For AKM and UKM the three parameters of the kernels $(\sigma^2, \theta, \tau^2)$ are obtained using maximum likelihood estimation~\cite{Rasmussen2006,Santner2003}. To asses the quality of the obtained metamodels, the predictivity coefficient $Q_2$ is computed on a test sample of $n_t=1000$ points uniformly distributed over $[0,1]^d$:
\begin{equation}
Q_2(y,\hat{y}) = 1 - \frac{\sum_{i=1}^{n_t}(y_i-\hat{y}_i)^2}{\sum_{i=1}^{n}(y_i-\bar{y})^2}
\label{eq:Q2}
\end{equation}
where $y$ is the vector of the values at the test points, $\hat{y}$ is the vector of predicted values and $\bar{y}$ is the mean of $y$.

\begin{figure}[!ht]%
\begin{center}
\includegraphics[width=0.9\textwidth]{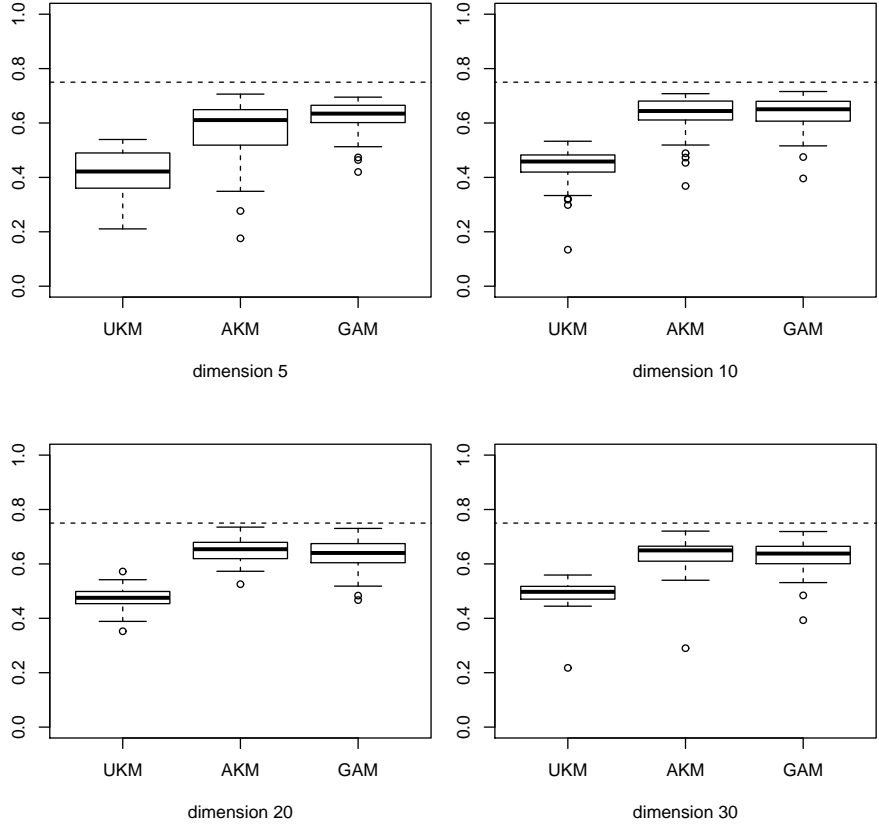}
\end{center}
\caption{Boxplots of the predictivity coefficients $Q_2$ for three emulators: Usual Kriging Model (UKM), Additive Kriging Model (AKM) and GAM. For a given boxplot, the variability is due to the choice of the DoE which is repeated 50 times.}%
\label{fig:BOXQ2}
\end{figure}

\noindent
As the parameter estimation accuracy and the overall quality of an emulator are likely to fluctuate with the DoE, we repeated 50 times each emulator's building and testing for various DoE. The results are presented in figure~\ref{fig:BOXQ2}. Conversely to what we observed in section~\ref{sec:khd}, the predictivity of the Kriging model based on a separable kernel does not fall to zero when the dimension increases. As we impose the additive part of $g$ to explain $75 \%$ of its variance, the value of the coefficient $a_1$ is increasing with $d$ and the g-function becomes smoother. As a result, the range parameter $\theta$ increases with $d$ (we have $\theta \approx 0.5$ for $d=5$ and $\theta \approx 2$ for $d=30$) so the predictivity of the models based on separable kernels do not fall to zero as previously.

\begin{figure}[ht]
\centering
\subfigure[$d=10$]{
\includegraphics[width=4cm]{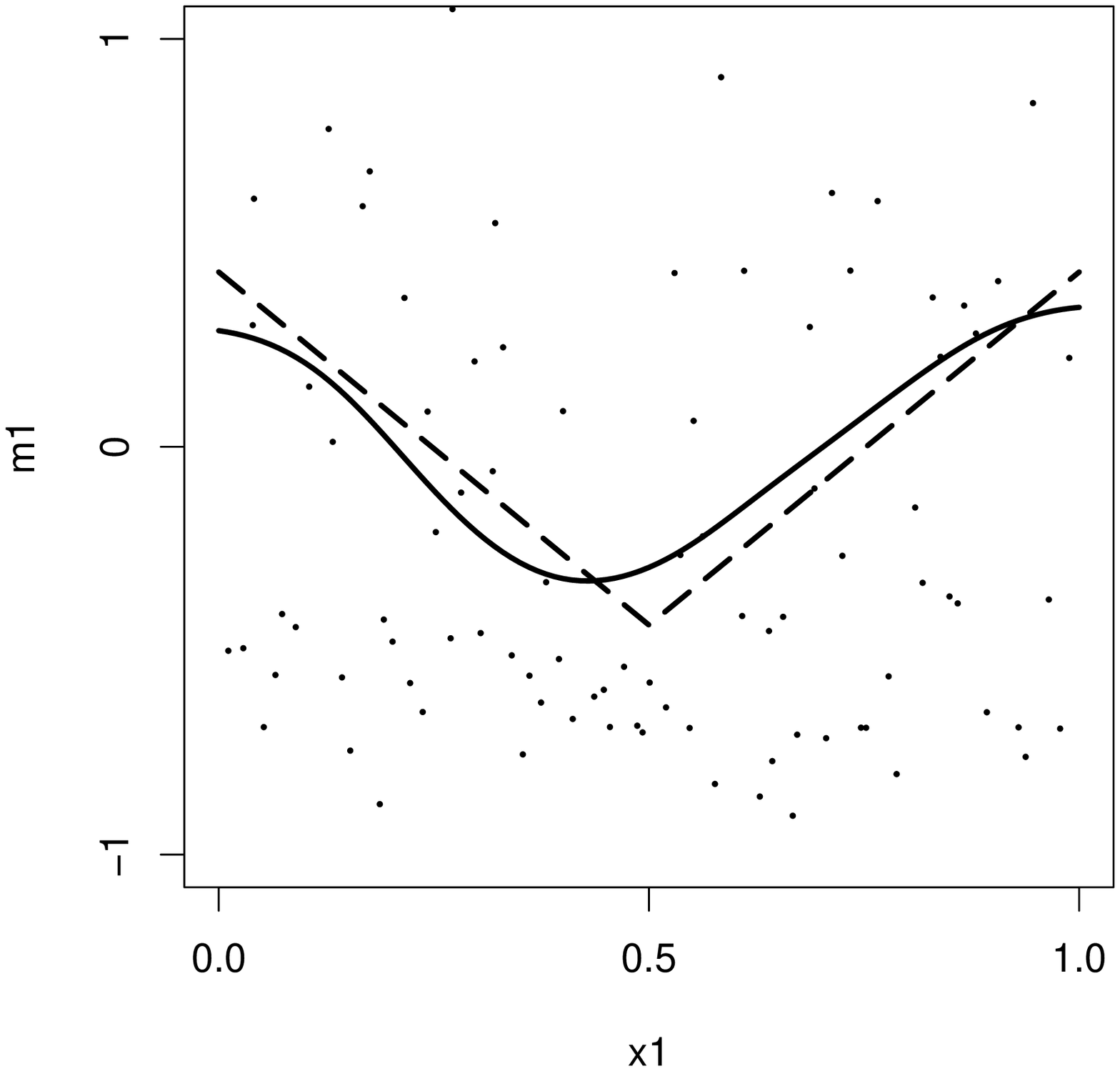}
\label{fig:subfig1}
}
\subfigure[$d=30$]{
\includegraphics[width=4cm]{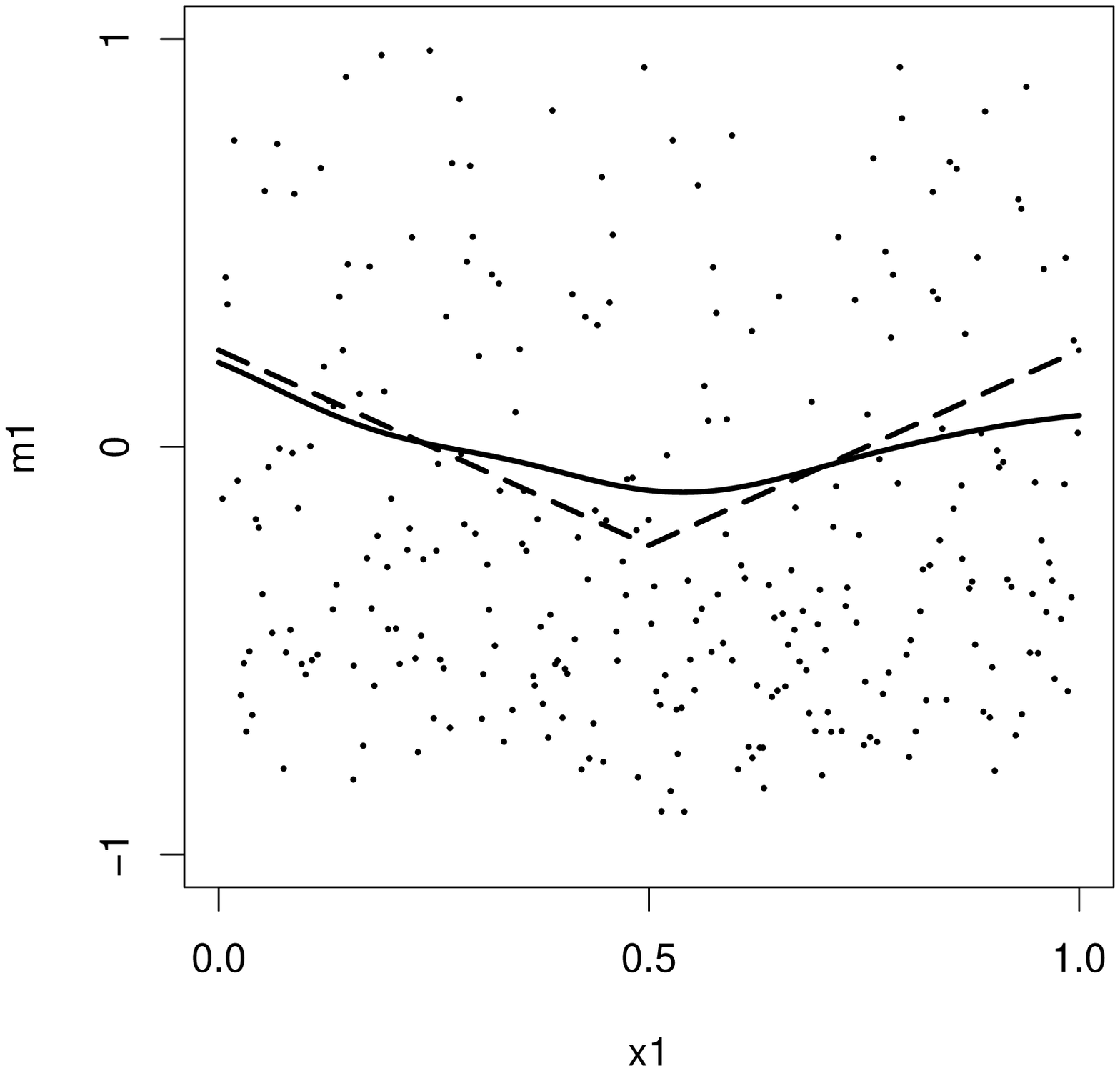}
\label{fig:subfig2}
}
\subfigure[$d=50$]{
\includegraphics[width=4cm]{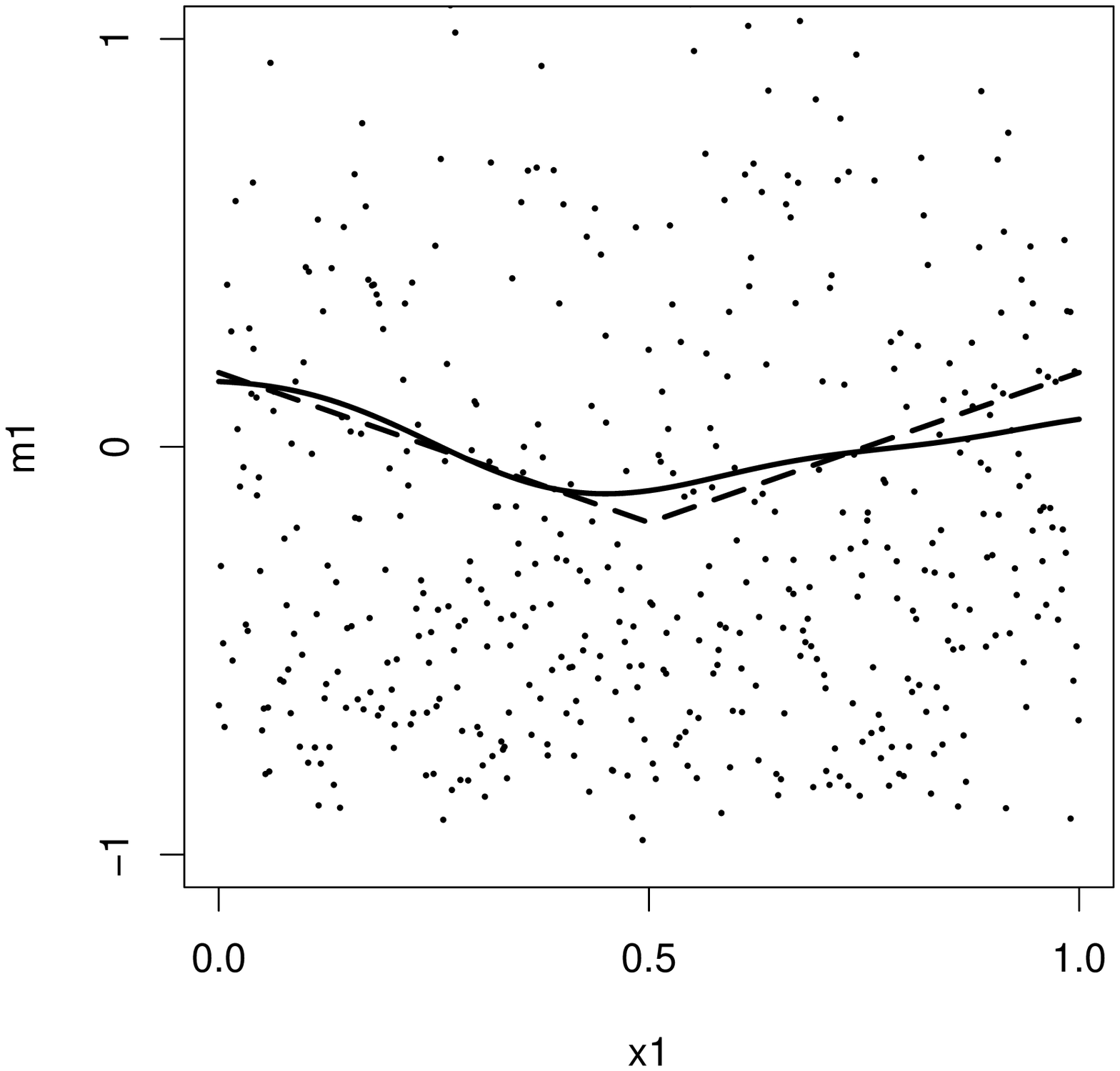}
\label{fig:subfig3}
}
\caption{Representation of the univariate submodels $\tilde{m}_1(x_1)$ (solid lines) for three additive Kriging models. As a comparison, the analytical main effects are given by the dashed lines. The bullets denote the centered observation points.}%
\label{fig:dim4dir}
\end{figure}

\medskip
\noindent
In order to illustrate the increasing smoothness of $g$, we represent the univariate submodels $\tilde{m}_1$ for various values of $d$ (fig.~\ref{fig:dim4dir}).  Even if the observation points do not show any obvious trend, the submodels are close to the analytical main effects.

\section{Concluding remarks}
The proposed methodology seems to be a good challenger for additive modeling. On the first example, additive models appears to be well suited for high-dimensional modeling with a DoE budget of $10 \times d$ whereas Kriging models based on standard kernels fail to recover the function to approximate. One important result is that additive kriging models succeed to extract the additive trend of the function to approximate even if this function is not purely additive. 

\medskip
\noindent
The proposed  additive models take advantage of additivity, while taking advantage from GP features. For the first point we can cite the complexity reduction and the interpretability of additive models. For the second, the main asset is that GP models include a prediction variance for the model but also for each submodel. This justifies the fact of modeling an additive function on $\mathds{R}^d$ instead of building $d$ metamodels over $\mathds{R}$ since the prediction variance is not additive. At the end, the proposed methodology is fully compatible with Kriging-based methods and its versatile applications. For example, one can choose a well suited kernel for the function to approximate or use additive Kriging for high-dimensional optimization strategies relying on the expecting improvement criteria.

\medskip
\noindent
In this article, we only considered isotropic kernels. As for separable kernel, the use of additive kernels can easily be extended to anisotropic kernels (ie one range parameter $\theta_i$ per direction) but additive kernels also allow to define one variance parameter $\sigma^2_i$ per direction. This feature, which is not possible for separable kernels, can enable additive models to approximate functions for which the variance depends on the direction. However, the total number of parameters would be $2d+1$ and the practicability of their estimation deserves to be studied in detail.

\bibliographystyle{plain}
\bibliography{reference}

\begin{thebibliography}{10}

\bibitem{Buja1989}
A.~Buja, T.~Hastie, and R.~Tibshirani.
\newblock Linear smoothers and additive models.
\newblock {\em The Annals of Statistics}, pages 453--510, 1989.

\bibitem{chiles1999geostatistics}
J.P. Chil\`es and P.~Delfiner.
\newblock {\em Geostatistics: modeling spatial uncertainty}, volume 344.
\newblock Wiley-Interscience, 1999.

\bibitem{Cressie1993}
N.~Cressie.
\newblock Statistics for spatial data.
\newblock {\em Terra Nova}, 4(5):613--617, 1992.

\bibitem{K-T.Fang2006}
K.~Fang.
\newblock {\em Design and modeling for computer experiments}, volume~6.
\newblock CRC Press, 2006.

\bibitem{fortet1985operateurs}
R.M. Fortet.
\newblock Les operateurs integraux dont le noyau est une covariance.
\newblock {\em Trabajos de estad{\'\i}stica y de investigaci{\'o}n operativa},
  36(3):133--144, 1985.

\bibitem{gaetan2009spatial}
C.~Gaetan and X.~Guyon.
\newblock {\em Spatial statistics and modeling}.
\newblock Springer Verlag, 2009.

\bibitem{Ginsbourger2009}
D.~Ginsbourger, D.~Dupuy, A.~Badea, L.~Carraro, and O.~Roustant.
\newblock A note on the choice and the estimation of kriging models for the
  analysis of deterministic computer experiments.
\newblock {\em Applied Stochastic Models in Business and Industry},
  25(2):115--131, 2009.

\bibitem{Hastie2011}
T.~Hastie.
\newblock {\em gam: Generalized Additive Models}, 2011.
\newblock R package version 1.04.1.

\bibitem{Hastie1990}
T.J. Hastie and R.J. Tibshirani.
\newblock {\em Generalized additive models}.
\newblock Chapman \& Hall/CRC, 1990.

\bibitem{loeppky2009choosing}
J.L. Loeppky, J.~Sacks, and W.J. Welch.
\newblock Choosing the sample size of a computer experiment: A practical guide.
\newblock {\em Technometrics}, 51(4):366--376, 2009.

\bibitem{Muhl}
T.~Muehlenstaedt, O.~Roustant, L.~Carraro, and S.~Kuhnt.
\newblock Data-driven {Kriging models based on FANOVA}-decomposition.
\newblock {\em to appear in Statistics and Computing}.

\bibitem{newey1994kernel}
W.K. Newey.
\newblock Kernel estimation of partial means and a general variance estimator.
\newblock {\em Econometric Theory}, 10(02):1--21, 1994.

\bibitem{Rasmussen2006}
C.E. Rasmussen and C.K.I. Williams.
\newblock Gaussian processes for machine learning.
\newblock 2005.

\bibitem{Roustant2009}
Olivier Roustant, David Ginsbourger, and Yves Deville.
\newblock {\em DiceKriging: Kriging methods for computer experiments}, 2011.
\newblock R package version 1.3.

\bibitem{Saltelli2000}
A.~Saltelli, K.~Chan, E.M. Scott, et~al.
\newblock {\em Sensitivity analysis}, volume 134.
\newblock Wiley New York, 2000.

\bibitem{Santner2003}
T.J. Santner, B.J. Williams, and W.~Notz.
\newblock {\em The design and analysis of computer experiments}.
\newblock Springer Verlag, 2003.

\bibitem{stone1985additive}
C.J. Stone.
\newblock Additive regression and other nonparametric models.
\newblock {\em The annals of Statistics}, pages 689--705, 1985.

\bibitem{RR}
R~Team.
\newblock R: A language and environment for statistical computing.
\newblock {\em R Foundation for Statistical Computing Vienna Austria ISBN},
  3(10), 2008.

\end{thebibliography}

\appendix

\section*{Appendix A: Proof of proposition \ref{addproc} for $d=2$}
Let $Z$ be a centered random process indexed by $\mathds{R}^2$ with covariance kernel $K(x,y)= K_1(x_1,y_1) + K_2(x_2,y_2)$, and $Z_T$ the random process defined by $Z_T(x_1,x_2)=Z(x_1,0)+Z(0,x_2)-Z(0,0)$. By construction, the paths of $Z_T$ are additive functions. In order to show the additivity of the paths of $Z$, we will show that $\forall x \in \mathds{R}^2$, $\p(Z(x)=Z_T(x))=1$. For the sake of simplicity, the three terms of $\V[Z(x)-Z_T(x)]=\V[Z(x)]+\V[Z_T(x)]-2 \C[Z(x),Z_T(x)]$ are studied separately:
\begin{equation*}
\V[Z(x)]=K(x,x)
\end{equation*}
\begin{equation*}
\begin{split}
\V[Z_T(x)] & = \V[Z(x_1,0)+Z(0,x_2)-Z(0,0)] \\
& = \V[Z(x_1,0)] + \V[Z(0,x_2)] + 2 \C[Z(x_1,0),Z(0,x_2)] \\
& \qquad + \V[Z(0,0)] - 2 \C[Z(x_1,0),Z(0,0)] - 2 \C[Z(0,x_2),Z(0,0)] \\
& = K_1(x_1,x_1) + K_2(0,0) + K_1(0,0) + K_2(x_2,x_2) + K(0,0) \\
& \qquad + 2 \left( K_1(x_1,0) + K_2(0,x_2)\right) - 2 \left( K_1(x_1,0) + K_2(0,0) \right) \\
& \qquad - 2 \left( K_1(0,0) + K_2(x_2,0) \right) \\
& = K_1(x_1,x_1) + K_2(x_2,x_2) = K(x,x)
\end{split}
\end{equation*}
\begin{equation*}
\begin{split}
\C[Z(x),Z_T(x)] & = \C[Z(x_1,x_2),Z(x_1,0)+Z(0,x_2)-Z(0,0)] \\
& = K_1(x_1,x_1) + K_2(x_2,0) + K_1(x_1,0) + K_2(x_2,x_2) \\
& \qquad  - K_1(x_1,0) - K_2(x_2,0) \\
& = K_1(x_1,x_1) + K_2(x_2,x_2) = K(x,x)
\end{split}
\end{equation*}
Those three equations implies that $\V[Z(x)-Z_T(x)]=0$, $\forall{x} \in \mathds{R}^2$. As $\E[Z(x)-Z_T(x)]=0$, we have $\p(Z(x)=Z_T(x))=1$ so there exists a modification of $Z$ with additive paths.

\section*{Appendix B: Calculation of the prediction variance}
Let consider a DoE composed of the 3 points $\{x^{(1)}\ x^{(2)}\ x^{(3)}\}$ represented on the left pannel of figure~\ref{fig:planprob}. We want here to show that although $x^{(4)}$ does not belongs to the DoE we have $v(x^{(4)})=0$.
\begin{eqnarray*}
v(x^{(4)}) & = & K(x^{(4)},x^{(4)}) - k(x^{(4)})^T \mathrm{K}^{-1} k(x^{(4)}) \\
& = & K(x^{(4)},x^{(4)}) - (k(x^{(2)})+k(x^{(3)})-k(x^{(1)}))^T \mathrm{K}^{-1} k(x^{(4)}) \\
& = & K_1(x^{(4)}_1,x^{(4)}_1) + K_2(x^{(4)}_2,x^{(4)}_2) - \\
& & \quad(-1\ \ 1\ \ 1) 
\begin{pmatrix}
K_1(x^{(1)}_1,x^{(4)}_1)+K_2(x^{(1)}_2,x^{(4)}_2) \\
K_1(x^{(2)}_1,x^{(4)}_1)+K_2(x^{(2)}_2,x^{(4)}_2) \\
K_1(x^{(3)}_1,x^{(4)}_1)+K_2(x^{(3)}_2,x^{(4)}_2)
\end{pmatrix}
 \\
& = & K_1(x^{(2)}_1,x^{(2)}_1) + K_2(x^{(3)}_2,x^{(3)}_2) - K_1(x^{(2)}_1,x^{(2)}_1) - K_2(x^{(3)}_2,x^{(3)}_2) \\
& = & 0
\end{eqnarray*}

\section*{Appendix C: Calculation of $\tilde{v}_i$}
We want here to calculate the variance of $Z_i(x_i)-\int{Z_i(s_i)\mathrm{d} s_i}$ conditionally to the observations $Y$.
\begin{equation*}
\begin{split}
\tilde{v}_i(x_i)&=\V \left[ \left.Z_i(x_i)-\int{Z_i(s_i)\mathrm{d} s_i} \right| Z(X)=Y  \right] \\
&=\V \left[ \left.Z_i(x_i) \right| Z(X)=Y  \right] - 2\C \left[ \left.Z_i(x_i),\int{Z_i(s_i)\mathrm{d} s_i} \right| Z(X)=Y  \right] \\
& \hspace{5cm} + \V \left[ \left.\int{Z_i(s_i)\mathrm{d} s_i} \right| Z(X)=Y  \right] \\
&= v_i(x_i) - 2 \left( \int{K_i(x_i,s_i)\mathrm{d} s_i} - \int{k_i(x_i)^T K^{-1} k_i(s_i) \mathrm{d} s_i} \right) \\
& \hspace{2cm} + \iint{K_i(s_i,t_i)\mathrm{d} s_i \mathrm{d} t_i} - \iint{k_i(t_i)^T K^{-1} k_i(s_i) \mathrm{d} s_i  \mathrm{d} t_i}.
\end{split}
\end{equation*}

\end{document}